\documentclass[10pt,letterpaper]{article}

\usepackage{cogsci}

\cogscifinalcopy % Uncomment this line for the final submission 

\usepackage{graphicx} %for includegraphics
\usepackage{caption}
\usepackage{threeparttable}  %for apa style tables
\usepackage{booktabs}

\usepackage{url} % so urls are nicely formatted
\usepackage{microtype} %further customise urls so the letter spacing is reduced
\let\oldurl\url
\renewcommand{\url}[1]{\texttt{\textls[-100]{\oldurl{#1}}}}

\usepackage{pslatex}
\usepackage{apacite}
\usepackage{float} % Roger Levy added this and changed figure/table
\usepackage{color,soul}
\usepackage{todonotes}
\usepackage{pdflscape}
\usepackage{subcaption}
\usepackage{makecell}
\title{The effect of diversity on group decision-making}
 
 \author{{\large \bf Georgi Karadzhov (gmk34@cam.ac.uk)} \\
   Department of Computer Science and Technology, University of Cambridge \\
   Cambridge, UK
  \AND {\large \bf Andreas Vlachos (av308@cam.ac.uk)} \\
   Department of Computer Science and Technology, University of Cambridge \\
   Cambridge, UK
   \AND {\large \bf Tom Stafford (t.stafford@sheffield.ac.uk)} \\
   Department of Psychology, University of Sheffield \\
   Sheffield, UK}

\begin{document}

\maketitle

\begin{abstract}
We explore different aspects of cognitive diversity and its effect on the success of group deliberation. To evaluate this, we use 500 dialogues from small, online groups discussing the Wason Card Selection task - the DeliData corpus. Leveraging the corpus, we perform quantitative analysis evaluating three different measures of cognitive diversity. First, we analyse the effect of group size as a proxy measure for diversity. Second, we evaluate the effect of the size of the initial idea pool. Finally, we look into the content of the discussion by analysing discussed solutions, discussion patterns, and how conversational probing can improve those characteristics.
\par
Despite the reputation of groups for compounding bias, we show that small groups can, through dialogue, overcome intuitive biases and improve individual decision-making. Across a large sample and different operationalisations, we consistently find that greater cognitive diversity is associated with more successful group deliberation. 
\par
Code and data used for the analysis are available in the repository: \url{https://github.com/gkaradzhov/cognitive-diversity-groups-cogsci24}.

\end{abstract}

\section{Introduction}

Individuals come together to discuss, analyse, and evaluate different points of view. Such group deliberation is considered an essential tool for tackling complex tasks. It is a cornerstone of decision-making in domains such as research, government and business. In academia, we have the peer review process \cite{peerreview2015,marcoci2022reimagining}, where scientists discuss and improve scientific reports and academic studies. In government, collective bargaining, resource allocation and community planning are driven by committees, task forces, and city councils, where groups of people try to come up with fair decisions and policies. In business, decisions are made again in small groups, such as boards of directors or management teams, responsible for running companies, budgeting, and hiring. The widespread deployment of group deliberation as a decision-making strategy makes it important to understand the forces which support or undermine good deliberation. The current work presents a data-driven approach to understanding one such key factor: cognitive diversity. %thought i should make explciit the distinction from demographic diversity (e.g. gender, race) which we do not speak to directly

Group deliberation is the dialogue %between a group 
that facilitates the careful weighing of information, providing adequate speaking opportunities, and bridging the differences among participants’ diverse ways of speaking and knowing \cite{burkhalter2002conceptual}. \citeA{mercier2011humans} review evidence that, in the right conditions, groups can outperform even the most capable %successful
individual members in their decision-making. These conditions include common goals among group members and the exchange of arguments. As well as groups exceeding the capacity of any individual member, another relevant standard is the decision which would be reached by simple aggregation or averaging of independent individual judgements \cite<the so-called ``Wisdom of the Crowd" effect,>{galton1907}. In the study of \citeA{navajas2018aggregated} small groups outperform the wisdom of the crowd standard. \citeA{navajas2018aggregated} argue that small groups can be more effective than both individuals of superior ability, and large crowds. Further, this and other evidence \cite{mercier2016, trouche2014arguments} supports the idea that the exchange of arguments is essential to successful group decision-making.

\par

That said, group decision-making is not universally successful. A tradition in social psychology has emphasised the potential for negative outcomes. A notable diagnosis is the perils of ``Group Think" \cite{Janis1972VictimsOG}. Potential mechanisms of failures of group decision-making include social loafing, herding, deference to others' preferences --- real \cite{mojzisch2010knowing}, or imagined \cite{prentice1996pluralistic} ---, failure to share knowledge available to individual group members \cite{Stasser1985PoolingOU}, underweighting dissenting voices, and amplification of individual cognitive biases. We note that the mechanisms supporting the Wisdom of Crowds effect relies on individuals having independent access to some true information (signal), and uncorrelated errors (noise). To the extent that individuals share systematic error, such as a shared cognitive bias, aggregation of individual judgements will reinforce error, not eliminate it. 

% More formally, this is also known as the Condorcet's jury theorem \cite<>{cjt-conderset-de1785essai}. Under the assumption that all voters in a group have independent probabilities to vote a correct decision and that probability is $>$ 0.5. Thus, by increasing the size of the group to vote, the probability of reaching a correct solution approaches 1. 

% % other topics
%\subsection{Online reasoning}

% Review and include?
% \cite{matsuyama2015four} argue that decision-making is driven by good conversational dynamics and equal sharing of ideas. On the other hand,
% \cite{stafford2018confronting} argue that individuals' cognitive biases, can affect the argument content and the conversational dynamics of the group, ultimately increasing the chances of a wrong answer. 
% Thus combating biases was explored as a way to improve group decision making by \cite{rozenblit2002misunderstood}, focusing on the \textit{illusion of exploratory depth}, and   \cite{lord1984considering} who suggested counter-arguments as constructive elements of a discussion.
% One could attribute better group performance to the individual intelligence of one or more participants, but \cite{geil1998collaborative} and \cite{schulz2006group} challenge this premise and argue that other factors such as deliberation, group dynamics, and diversity play a bigger role.

To study group decision-making under conditions of bias, we use the Wason Card Selection Task \cite{wason1968reasoning}, an iconic task used to study human reason ~\cite{evans2016reasoning}. For our purposes, it has two important qualities. First, with high reliability, the majority of people get the task wrong, and they fail in a consistent way. This suggests the presence of a confirmation, or matching, bias which draws people to an intuitive but incorrect response. This makes the task an attractive paradigm for investigating how group function may interact with individual biases. Second, the Wason task has a correct answer which can, once understood, be relatively easily explained and understood by others. This underlies the remarkable prior result that while individuals tend to fail on the Wason Task, small groups tend to succeed  \cite<discussed in>{mercier2011humans}.

\subsection{Cognitive diversity in group decisions}

\par
An established finding in studies of group decision-making is the potential benefits of diversity \cite{bang2017making, davis2014crowd, page2008difference}. Diversity can be in the form of differences in the backgrounds, knowledge or cognitive styles that individuals bring to a group, as well as the different roles they may adopt or be assigned within a group. In the limit, too extreme diversity can hamper group function --- particularly where it facilitates intragroup conflict or hinders agreement on common goals --- but otherwise, there are good reasons to expect greater diversity to enhance group decision-making.

Diversity can be improved in a number of ways. One approach is by differing demographic or professional characteristics ~\cite{wang2011diversity} of the participants. This can also be in part achieved by considering groups of larger sizes - i.e.\ by having a larger group, there is an increased chance that the group would have people from different backgrounds and experiences. An alternative way is to adopt procedures or structures which facilitate expressing and exchanging diverse ideas \cite{dalkey1963experimental}. Our focus here is on \textit{cognitive} diversity, which reflects differences between individuals in knowledge, processes or actions within the group context and is distinct from (but may stem from) demographic differences, such as gender or ethnic diversity.

%One approach to improve group deliberation is by introducing diversity within a group early by altering the participant selection to include participants from various backgrounds \cite{wang2011diversity}. If we consider randomly distributed groups, \cite{schulz2006group} show that the diversity of participants' solutions benefits the group's final performance and is a mechanism to fight confirmation bias within the group. Going one step further, \cite{barrera2023wisdom} argue that group decision-making can be improved by exploiting the anchoring effect and that having people with extreme opposing views is beneficial for the overall performance of the group.

\subsubsection{The current work, hypotheses.}
In this work, we leverage a unique dataset of group decisions, the DeliData corpus \cite<>{karadzhov2023delidata}. In DeliData, 500 groups provided initial judgments and then discussed 
%a classic reasoning problem - 
the Wason Card Selection Task, for which full transcripts were recorded. Individuals then submitted their updated answers. This dataset allows us to look, with a high degree of granularity, at the features of groups and the deliberative processes which generated successful and unsuccessful deliberation. Furthermore, DeliData contains data indicating participants' solutions, arguments, and deliberative markers, which would be beneficial for rich and multifaceted analysis. 

Concretely, we look into three aspects of cognitive diversity - (i) indirectly, via the proxy of group size, (ii) via the distribution of initial solutions possessed by the group, and (iii) directly, via the content of the discussions, the solutions considered by the group, as well as deliberation markers affecting said content.

\par
Given previous research, as well as the three aspects of cognitive diversity, we will be testing the following hypotheses. First, that larger groups will demonstrate better performance. This has been shown in other studies \cite{navajas2018aggregated, Hill1982GroupVI}, but not --- to our knowledge --- in the Wason Task, nor for other tasks which elicit strong individual biases (for which groups may function to aggregate rather than eliminate)  as a characteristic that may affect group performance.

Second, we hypothesise that a greater diversity of initial ideas will support better group performance. Following \citeA{schulz2006group} we index this by inspecting the number of distinct individual solutions submitted by participants prior to entering the group discussion. 
Third and finally, that diversity in discussion content will support better group performance. This was shown for forecasting by \citeA{wintle2023predicting}. We index this based on the distinct solutions mentioned in the conversation, as well as the transitions between mentioned solutions. We further hypothesise, that conversational probing (i.e. asking questions that provoke further discussion \cite{karadzhov2023delidata}) has an effect over the diversity of discussion content. The premise is that conversational probing can be considered as a possible intervention that would increase the content and cognitive diversity of a discussion.

\section{Data \& Methods} 

\subsection{DeliData}

\begin{table}[]
    \centering
    \begin{tabular}{|l|r|}
        \hline
        \# of dialogues & 500  \\ \hline
        \# of utterances & 14003 \\ \hline
        AVG \# of utterances / dialogue & 28 \\ \hline
        AVG group size & 3.16 \\ \hline
        AVG utterance length & 8.59 \\ \hline
        \# of Probing utterances & 1739 \\ \hline
        Solo Performance & 0.59 \\ \hline 
        Group performance & 0.72 \\ \hline 
        Performance Gain & 0.13 \\ \hline 
    \end{tabular}
    \caption{Summary statistics for the DeliData corpus}
    \label{tab:delidata_stats}
\end{table}

The DeliData corpus is described in \citeA{karadzhov2023delidata}. It comprises 500 small group discussions of the Wason Task, which vary both in terms of group composition (2 to 5 participants) and performance (both high and low, and pre- (solo) and post-discussion (final), performance). Statistics of the DeliData corpus are presented in Table~\ref{tab:delidata_stats}. 

To create the dataset, the Wason Card Selection Task was presented in its traditional form: it is explained that cards have a number on one side and a letter on the other.  Four cards are shown displaying an even number (`E'), odd number (`O'), consonant (`C') and vowel (`V'). The task is to answer the question ``Which card(s) should you turn to test the rule: \textbf{All cards with vowels on one side, have an even number on the other}". The intuitive but wrong answer (henceforth the Lure) is to turn the even number and vowel (`EV'). The correct answer (henceforth the Target) is to turn over the odd number and the vowel (`OV').

\citeA{karadzhov2023delidata} report that 64\% of the groups performed better following the discussion compared to their initial performances. In 44\% of the groups who had at least one correct answer as their final solution, none of the participants had begun with the solution. This demonstrates the benefits of group discussion and suggests that more is occurring than mere propagation of the best answers among the groups. 

Moreover, \citeA{karadzhov2023delidata} provides utterance-level annotations that mark deliberative and argumentation cues. The full details of the annotation scheme and corpus are presented in \citeA{karadzhov2023delidata}, however in the context of this work, we are interested in the phenomenon of \textit{conversational probing}. Conversational probing is defined as any utterance that provokes discussion, deliberation or argumentation. This is further split into three categories - Moderation (i.e.\ modelling group dynamics), Reasoning (probing further argumentation), and Solution (i.e.\ enquiring about a specific solution).

\subsection{Analytic strategy}

The Wason Card Selection task has one correct answer, allowing for a direct and objective measure of performance. In this work, we leverage the fine-grained scoring from \citeA{karadzhov2023delidata}, where 0.25 points each are given for i) selecting either card in the Target (O and V)  and (ii) for \textbf{not} turning unnecessary cards (E and C). Therefore, if the participant submitted a correct solution, their score would be 1, if they are off by one card - 0.75 and so on. If participants selected the Lure (EV), their fine-grained score would be 0.5. Similarly, selecting all cards (CEOV) or none, will have a fine-grained score of 0.5.

\par
\textbf{Performance gain} is calculated by subtracting the \textbf{task performance} of the solo solution from the task performance in the final submission. For example, if a participant's solo submission was 0.5 and improved to 0.75 after the discussion, the performance gain would be $0.75 - 0.5 = 0.25$. In this work, we investigate \textbf{group performance gain} which is the averaged performance gain of each of the participants.

\subsubsection{Different measures of diversity}
\label{sec:measure-diversity}
% --- for consistency (in a longer paper?) there would be material here about defining the measures of diversity

First, we investigate group size as a proxy measure of diversity, and can be compared to a baseline of pre-discussion, solo performance (groups of size 1). The premise is that more members participating in a dialogue would naturally increase cognitive diversity.
Following, we consider the diversity of initial ideas, which is calculated by counting the number of unique solo solutions submitted by the group's participants. 
Finally, we analyse the factors measuring the diversity of discussed solutions and their contribution towards the performance gain. To this end, we evaluate the following characteristics of a diverse discussion. 

\begin{itemize}
    \item \textbf{dialogue length} measured by the number of utterances. The premise is that longer conversations will naturally contain more mentions of solutions, thus contributing towards a more diverse discussion:
    
    \item Number of \textbf{unique solutions} that are mentioned in a discussion. We hypothesise, that conversations with positive performance gain will contain more mentions of unique solutions, indicating that participants were actively discussing a diverse set of possible solutions to the Wason task.
    
    \item We analyse how participants discuss solutions. To measure this, we extract the solutions participants talk about in every utterance by keeping track of Wason card mentions \footnote{And the two special cases when participants mentioned ``none'' or ``all'' where we consider that their solution is to select none or all of the cards}. Then, we calculate the occurrences of all transitions between solutions in the form of a from-through-to transition triple.  For example, if the conversation starts with the lure (EV), then transitions to a solution that turns every card, and then to the correct solution, the transition triplet would be ``EV-CEOV-OV''. Having these transition triplets, we measure the following three quantities:
    \begin{itemize}
        \item \textbf{Number of unique transitions}
        \item \textbf{Number of stuck transitions} - where participants recalled the same solution within a triplet, for example, EV-EV-EV.
        \item \textbf{Number of circular transitions} - where participants mention one solution, then an interim solution and finally return to the first solution. In our notation, this would be represented as: OV-CEOV-OV, if participants first discuss odd number + vowel, then discuss turning in all cards, and then they return back to odd + vowel. This is a special case of stuck solutions, where participants may try to ``break the cycle'' but ultimately return to a previous solution.
    \end{itemize}
\end{itemize}

 We conclude this section by also analysing the interaction of the measures above and the probing utterances, as annotated in DeliData \cite<>{karadzhov2023delidata}

\section{Results}

\subsection{Effects of group size on diversity}

%\begin{figure}[H]
%    \includegraphics[scale=0.05]{scatter_performances.png}
%\caption{Performance of individuals (pre-discussion) and groups (average, %post-discussion) for groups of different sizes.}
%\end{figure}

    % \label{fig:sub1}
\begin{figure}[H]
\centering
\includegraphics[width=0.45\textwidth]{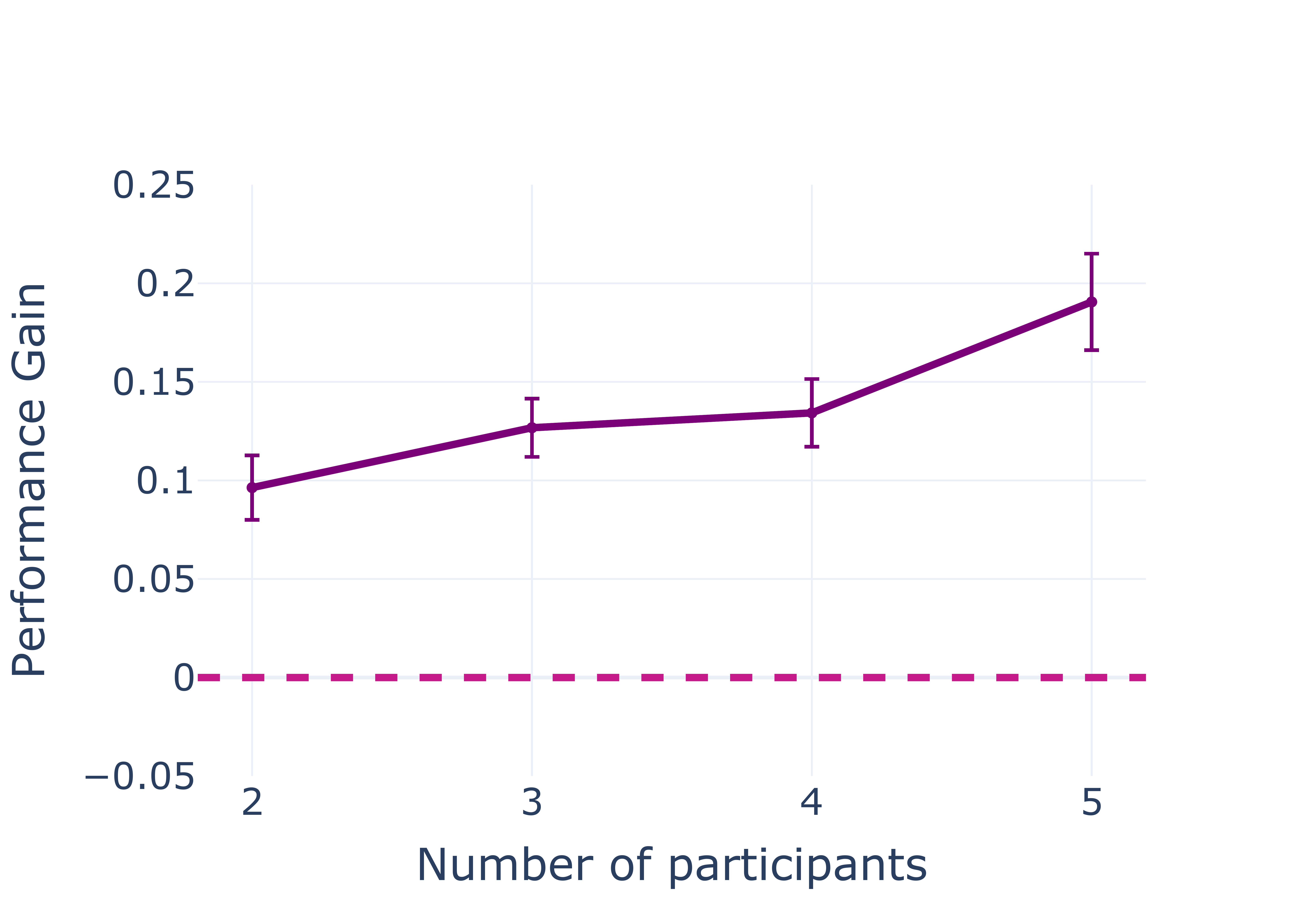}

\caption{Average performance gain against group size}
    % \label{fig:sub1}
\label{fig:analysis_groupsize_performances}
\end{figure}

%We compare the solo and group performances of different group sizes in Figure~\ref{fig:analysis_groupsize_performances}. We show substantial improvement between individuals solving the task by themselves and the same participants arranged into groups of as little as two participants (dyads). Following, we observe further improvement with the increase in group size. This improvement is represented by both increased final task performance, as well as increased performance gain.  This finding is consistent with \cite{Hill1982GroupVI}, who show that while the differences are small, group size is correlated with performance.  
As a first step of this analysis, we are hypothesising that group size is a factor contributing to better cognitive diversity and ultimately improved decision-making performance (as measured by performance gain).

Considering only the size of the group and the performance gain, we show that the performance gain of a group increases consistently with group size, as shown in Figure~\ref{fig:analysis_groupsize_performances}. To confirm the significance of these results, we perform a 1-sample t-test against the baseline performance gain of 0. This yields p-values of 2.5\textsuperscript{-8}, 3.5\textsuperscript{-15}, 4.1\textsuperscript{-12}, and 2.8\textsuperscript{-10} for groups of 2, 3, 4, 5 respectively. This indicates that for all group sizes, the performance gain was significantly above 0. 
Next, we performed an ANOVA \cite<>{girden1992anova} comparison between groups of sizes 2 to 5. This showed a significant result (p=0.0251), confirming that group size affects performance gain. Following this, we did a post-hoc analysis, performing a t-test between different group sizes. The results (also presented in Table~\ref{tab:groupsize_significance}) of the test show that there is no significant difference between each of the groups. This is also consistent and confirms the findings of \cite{Hill1982GroupVI}. That said, \citeA{Hill1982GroupVI} argues that groups (3+ people) perform better than dyads (dialogues). To this end, we also confirm this claim on the 500 dialogues of DeliData, showing that groups (size$>=$3) significantly outperform dyads (size=2) with a p-value of 0.03.  

\begin{table}[]
    \centering
    \begin{tabular}{|l|l|}
     \hline
        Groups & p-value  \\ \hline
        2 \& 3 & 0.17 \\ \hline
        3 \& 4 & 0.75 \\ \hline
        4 \& 5 & 0.06 \\ \hline
       \textbf{2 \& 3+} & \textbf{0.03} \\ \hline
    \end{tabular}
    \caption{Comparison between the effect of group size on performance gain. Significant values in \textbf{bold}.}
    \label{tab:groupsize_significance}
\end{table}

\begin{figure}[H]
\centering
\includegraphics[width=0.4\textwidth]{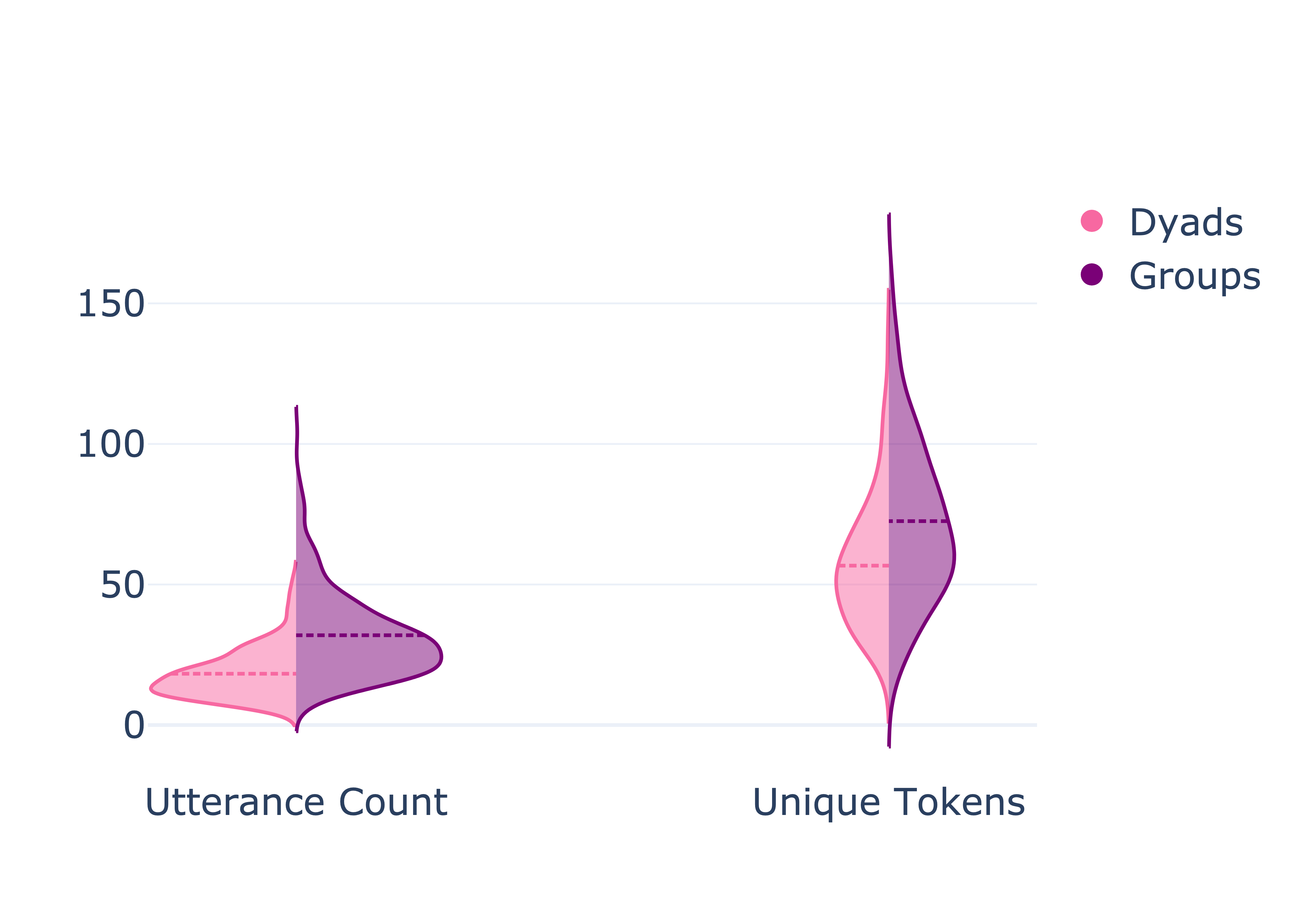}
% \caption{Comparison between performance gains of groups of various sizes.}
    % \label{fig:sub1}
\caption{Comparison between conversational statistics of two-party dialogues (dyads) and group dialogues (groups)}
\label{fig:dyadsvtriads}
\end{figure}

\par
Overall, we find that group size has an effect on performance gain. However, group size being a proxy measure, we cannot directly test whether the increase in performance gain is due to increased cognitive diversity or other factors. That said, in this work, we further investigate two characteristics that measure cognitive diversity directly. First, larger groups may benefit from an increased idea pool, which we investigate in detail in the next section - \textbf{Diversity of initial ideas}. Secondly, larger groups may engage in more diverse conversational strategies, thus incorporating more ideas and arguments during the group discussion. Figure~\ref{fig:dyadsvtriads} compares basic conversational statistics between dyads and groups. Dialogues between two interlocutors have mostly between 10 and 25 utterances, while group dialogues have a wider range of lengths, with a long tail of dialogues longer than 50 utterances. Likewise, groups in these dialogues tend to use a larger vocabulary, as shown on the violin plot of the unique tokens. We further evaluate conversational patterns and characteristics in section \textbf{Diversity of discussed solutions}. 

%\begin{figure}
% \centering
%    \includegraphics[scale=0.05]{violin_dyads_groups_performances.png}
%\caption{Comparison between task performances \\ of individuals and %groups of various sizes.}
%    \label{fig:sub1}
%\end{figure}%

%Besides conversation statistics, we analyse the difference in task performance. On the right-hand side of Figure~\ref{fig:dyadsvtriads}, we show violin plots of the solo performance, group performance and performance gain.  Verifying for the initial conditions first, the solo performance of both types of groups is comparable - 0.597 and 0.585 and are similarly distributed. On the other hand, the collective performance of these groups was 0.694 for two-party conversations and 0.724 for multi-party, thus, the performance gains are 0.096 and 0.139, respectively. These results are consistent with \cite{Hill1982GroupVI}, who showed that larger groups are better than dyads. We argue that multi-party (as opposed to two-party) discussions lead to more elaborate deliberation strategies, ultimately leading to increased problem-solving performance.  

\subsection{Diversity of initial ideas}
\label{subsec:ideas_group}

\begin{figure}[H]
\centering
\includegraphics[width=0.5\textwidth]{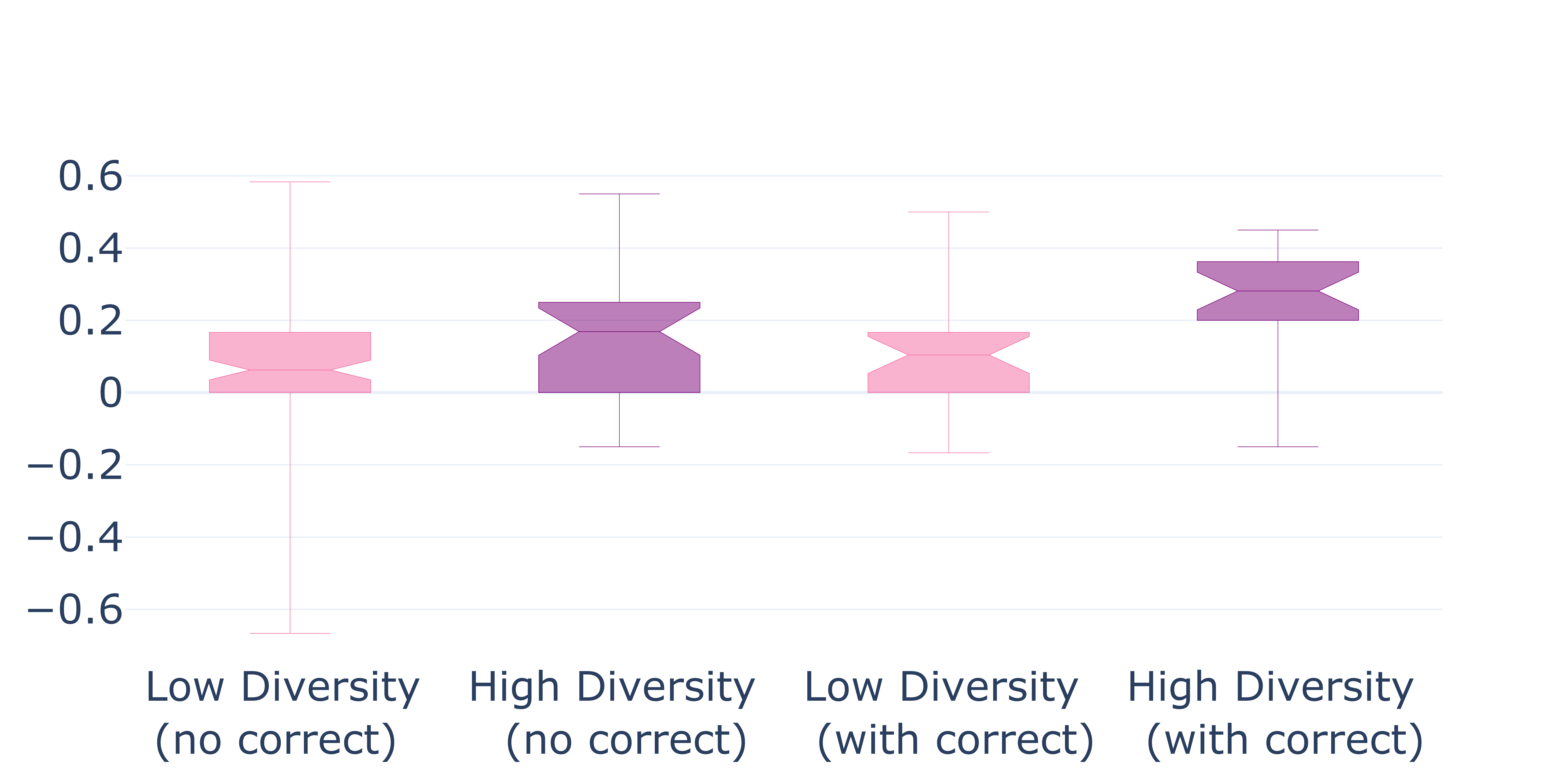}
\vspace*{0mm}
\caption{Performance gain of groups with high (purple) and low (pink) diversity of initial submissions. Groups are also split by whether any initial submissions contained a correct answer (right two) or did not contain any correct answers (left two).}
\label{fig:analysis_diversity_of_submissions}
\end{figure}

Herein, we investigate how the diversity of the initial pool of ideas (indicated by submitted solutions in participants' solo phase) affects performance gain. We hypothesise that more diverse initial ideas will provide a basis for more discussion and a more balanced starting point, ultimately leading to an improved conversation performance (i.e.\ increase in performance gain).
Figure~\ref{fig:analysis_diversity_of_submissions} shows how initial ideas, in terms of both diversity and the presence of a correct solution among group members, affect %post-discussion
performance gain. We compare two factors - (i) whether the initial idea pool contained a correct answer and (ii) the level of diversity (i.e.\ number of unique solutions) of the initial idea pool. We show that regardless of whether the initial idea pool contained a correct answer, high diversity is always beneficial for the group's performance gain.  Further, we analyse the effects of these two factors by fitting a linear regression, where performance gain is predicted based on the level of diversity and whether the initial idea pool contained a correct answer. The \textit{diversity} component has a statistically significant contribution with a p-value of 0.001. On the other hand, the p-value of the component indicating a correct initial solution is 0.079. 

\par
These results show that the diversity of initial ideas is a significant component of a constructive conversation. However, a reasonable question can be asked - \textit{how does the size of the group confound the diversity of the initial ideal pool with respect to performance gain?} Intuitively, larger groups would have more ideas, thus the chance of encountering a diverse pool of ideas is higher.
To evaluate this interaction between group size and the diversity of initial ideas, we predict performance gain using a linear regression model fitted with the least squares method. We consider three linear models with the following setups - we predict performance gain based on (i) group size, (ii) diversity of initial ideas, and (iii) a combination of (i) and (ii), i.e.\ group size + diversity of initial ideas. We then compare these models using ANOVA.

First, we compare models (i) and (iii), testing whether including diversity of ideas to group size gives us extra predictive power. The ANOVA comparison estimates that these two models are significantly different, with a p-value of 0.0006. Following, we compare the inverse - models (ii) and (iii), to test whether including group size to diversity of ideas gives us extra predictive power. This time, the difference is not significant, with a p-value of 0.81. 
\par 
This analysis shows that most of the predictive power comes from the diversity of initial solutions, and including the number of participants does not bring additional predictive power. Notably, another perspective would be that having a higher diversity of initial solutions already captures the information of having a larger group size.

\subsection{Diversity of discussed solutions}
\label{subsection:analysis:diversity_of_discussion}

%\begin{table}
%    \centering
%    \begin{tabular}{|l|c|c|}
%    \hline
%         & \textbf{p-value} & \textbf{coefficient } \\ \hline
%       \textbf{dialogue length}  & \textbf{0.004}  & \textbf{-0.235} \\ \hline
%       \textbf{unique solutions}  & \textbf{0}  & \textbf{0.316} \\ \hline
%       \textbf{unique transitions}  & 0.305 & 0.038 \\ \hline
%       \textbf{stuck transitions}  & 0.355  & 0.044 \\ \hline
%        \textbf{circular transitions} & \textbf{0.004} & \textbf{0.177}  \\ \hline
%    \end{tabular}
%    \caption{Results of fitted regression with different factors representing the diversity of solutions and predicting performance gain. %In bold, we show the statistically significant results.}
%    \label{tab:diversity-topics}
%\end{table}

\begin{table}
    \centering
    \begin{tabular}{|l|c|c|}
    \hline
         & \textbf{Correlation} & \textbf{PCR coef.} \\ \hline
       \textbf{dialogue length}  & \textbf{0.10} & \textbf{0.01} \\ \hline
       \textbf{unique solutions}  & \textbf{0.28} & \textbf{0.02} \\ \hline
       \textbf{unique transitions} & \textbf{0.22} & 0.02 \\ \hline
       \textbf{stuck transitions}  & 0.02 & -0.02  \\ \hline
        \textbf{circular transitions} & \textbf{0.19} & \textbf{0.01} \\ \hline
    \end{tabular}
    \caption{Results of correlation and PCR analyses between different factors representing the diversity of solutions and performance gain. In bold, we show the statistically significant values with p-value $<$ 0.05.}
    \label{tab:diversity-topics}
\end{table}

Here we use the five indicators outlined in the \textbf{Analytic strategy} subsection - dialogue length, number of unique solutions, number of unique transitions, stuck transitions and circular transitions. In order to evaluate these five indicators and their relationship to performance gain, we perform (i) a correlation study using \cite<>{pearsonR} and (ii) a regression analysis using Principal Component Regression (PCR) \cite<>{massy1965principal}. For the latter, we transform the input feature space using principal component analysis. This can handle adequately the multicollinearity present among the features in the dataset. We implement it by fitting a linear regression over the transformed feature space. We then transform the coefficients of the regression back to the original feature space to obtain the final coefficients.

The values of the correlation analysis and the PCR analysis are presented in Table~\ref{tab:diversity-topics}. If we consider the correlation statistic, out of the five factors, four are statistically significant in predicting the performance gain - the dialogue length, the number of unique discussed solutions, the number of unique transitions, and the number of circular transitions. Further, if we consider the correlation coefficients, the number of unique solutions (0.28) is considered the most important feature, followed by the number of unique transitions (0.22), and the circular transitions (0.19). Similarly, if we analyse the results of the PCR analysis, dialogue length, number of unique solutions, and circular transitions are significant. Similarly to the correlation analysis, the number of unique solutions has the highest importance, based on the coefficients from the principal component regression. 
\par
These results confirm the hypothesis that the diversity of discussed solutions correlates with group performance gain. Similarly to the diversity of the initial ideas, the number of unique solutions discussed is highly indicative of well-performing groups. Following similar intuition, we argue that this is both a measure of how diverse the discussion is, as well as it is a proxy for how open the group is towards new ideas.
\par
In the previous subsection (\textbf{Diversity of initial ideas}), we showed that groups with a high diversity of initial solutions achieve better performance gain. Likewise, in this section, we showed that the number of discussed solutions is also associated with increased performance gain. However, not all groups start with a diverse set of initial ideas. A reasonable research question would be - \textit{what happens in conversations where groups with low initial diversity achieve a high number of discussed ideas}. To evaluate this, we test the hypothesis that conversational probing affects dialogues with a low diversity of initial ideas. The premise is that if participants \textit{probe} the conversation more often, this will push the group to discuss novel ideas. To this end, we calculate a Pearson-R correlation of 0.34 between the number of conversational probes and the diversity of discussed solutions.

% \begin{figure}[H]
% \centering
% \includegraphics[width=0.5\textwidth]{probing_uniquesols.png}
% \vspace*{0mm}
% \caption{Scatter plot of relation between unique solutions discussed and number of probing utterances, for conversations with low diversity of initial ideas.}
% \label{fig:probing_to_uniquesolutions}
% \end{figure}

To further analyse this effect, we take advantage of the more granular \textit{probing} annotations in DeliData, namely whether the probing was related to \textbf{Moderation}, \textbf{Reasoning} behind a solution, or whether a specific \textbf{Solution} should be picked. To that end, we perform correlation and PCR studies to investigate the relationship between different kinds of probing and the diversity of discussed solutions. Results of this analysis are shown in Table~\ref{tab:probing-uniquesols}. We see that both probing for reasons and solutions are a significant contributor towards predicting the diversity of discussed solutions. The correlation coefficients show that probing for reasons (0.41) has a higher association with the diversity of discussed solutions than simply enquiring about solutions (PearsonR = 0.26). Likewise, the PCR coefficient for probing for reasons is 0.83, and probing for solutions is 0.43. This indicates that regardless of the type of analysis (correlational and predictive), we show the importance of probing for reasons, followed by probing for solutions. 
This is an interesting result, showing the importance of reasoning and also confirming the findings of \cite<>{mercier2011humans} about the value of deliberation. Moreover, this gives a basis for future work on interventional analysis and applications, where conversational probing can increase cognitive diversity and improve conversational outcomes. 

\begin{table}[H]
    \centering
    \begin{tabular}{|l|c|c|}
    \hline
         & \textbf{Correlation} & \textbf{PCR Coef.} \\ \hline
       \textbf{Moderation}  & -0.017  & -0.06 \\ \hline
       \textbf{Reasoning}  & \textbf{0.41}  & \textbf{0.83} \\ \hline
       \textbf{Solution}  & \textbf{0.26} & \textbf{0.43} \\ \hline
    \end{tabular}
    \caption{Results of correlation and PCR analyses of different types of probing and predicting the number of unique solutions discussed. In bold, we show the statistically significant values.}
    \label{tab:probing-uniquesols}
\end{table}

% What kind of transitions occur when they go from the wrong answer to the correct answer 

\section{Discussion}

Across three different operationalisations, we show that greater diversity is associated with enhanced benefits of discussion in small groups. 

On the chosen task, which has a high risk of individual bias and error, online discussion groups outperform individuals, and  groups of 3 and above outperform dyads. 

Groups incorporating more diverse individuals inherently have access to greater cognitive diversity. We find support for this idea more directly by showing that the diversity of initial solutions was a stronger predictor of group performance gain than the fact of any individual group member possessing the right answer.

Finally, at the greatest level of analytic granularity we deploy, we inspect the content of group discussion and show that groups which discuss a wider range of potential solutions are also more successful. The circularity and number of discussed solutions transitions were also predictive of an increase in performance gain, while stuck transitions were insignificant. Moreover, we show that conversational probing can increase the diversity of discussed solutions, even in groups which had a limited idea pool.

%We argue that in conversations with negative performance gain, participants are discussing similar solutions, thus reinforcing the \textit{confirmation bias}, which is associated with poor deliberation performance. On the other hand, groups that have good problem-solving performance exhibit a much more diverse set of discussion patterns, indicating individuals' ability to argue and to be open-minded towards different solutions.

\par
\subsection{Contributions}

This analysis of the DeliData corpus shows that it replicates the finding that group discussion of the Wason Task can reduce, rather than reinforce, the individual susceptibility to error on the task. It extends this finding to the increasingly relevant domain of online discussion among strangers. The finding is all the more remarkable for the low-commitment\footnote{In DeliData the participants were recruited from the crowd-sourcing website MechanicalTurk} 
 and the brief nature of the discussions (typically only 20 - 40 utterances).

By focusing on diversity, we confirm prior findings that the initial pool of ideas is an important factor in successful group deliberation \cite{schulz2006group}, as is the exchange of ideas during the discussion ~\cite{mercier2016}. We show that the association of diversity with successful deliberation extents beyond the mere content and volume of discussion to diversity in structure. Circularity in discussion is not necessarily negative --- it could represent focus and persistence which we also show is important predictor of performance improvement.

Additionally, we show how analysis of features of groups and group dialogue, captured at different levels of description, can be applied to the study of group decision-making. These approaches are applicable to other domains where dialogue supporting group decision-making is captured.

\subsection{Limitations}

Since our study is observational, not experimental, in nature, it is only possible to note associations, not causation. Ultimately, we suspect that successful groups generate diverse discussions, as well as vice versa. Our analysis supports the idea that diversity of membership, ideas and discussion will promote successful group decision-making, but decisive proof of this idea awaits testing by intervention studies. 

Further, in this study, we investigated the effect of cognitive diversity on performance gain. By definition, performance gain is affected by the baseline performance of the group - i.e.\ if a group has a low starting performance, the performance of this group can grow more compared to groups that start with high baseline performance. The vice-versa is also true - if a group starts with high (or even perfect) solo performance, during the discussion they can only decrease their performance leading to negative performance gain. We recognise that this is a limitation of the current study, as we only look at how much a conversation improved, disregarding the starting points. We performed a small-scale control experiment to evaluate the effect of this limitation. First, we consider the dialogues with high baseline performance, which we define as dialogues where more than 50\%  of the participants had the correct answer in their solo submissions. In DeliData, there are 9 dialogues out of 500 (1.8\%) where this is the case, suggesting that this affects a limited portion of the data. Second, we re-ran all of the analyses in this work without those 9 dialogues and confirmed that the removal of those does not affect the statistical significance tests, nor does it change any of the analyses performed as part of this work.

\subsection{Conclusion}

Despite the reputation of groups for compounding bias, these findings show that small groups can, through dialogue, enhance individual decision-making. Further, we offer a worked example of the potential for fine-grained analysis of dialogue to illuminate exactly how the benefits of deliberation come about. Across a large sample and different operationalisations, we consistently find that greater diversity is associated with more successful group deliberation.

\section{Acknowledgements}
 Georgi Karadzhov's PhD was supported by an EPSRC doctoral training scholarship. This research was additionally supported by a donation from Google.

\bibliographystyle{apacite}

\setlength{\bibleftmargin}{.125in}
\setlength{\bibindent}{-\bibleftmargin}

\bibliography{CogSci_Template}

\end{document}